\documentclass[letterpaper]{article} 
\usepackage{aaai2026}  
\usepackage{times}  
\usepackage{helvet}  
\usepackage{courier}  
\usepackage[hyphens]{url}  
\usepackage{graphicx} 
\usepackage{natbib}  
\usepackage{caption} 
\usepackage{amsmath}
\usepackage{amsthm}
\usepackage{booktabs} 
\usepackage{multirow}
\usepackage{amssymb}
\newtheorem{theorem}{Theorem}
\newtheorem{lemma}{Lemma}

\usepackage{algorithm}
\usepackage{algorithmic}

\usepackage{newfloat}
\usepackage{listings}
\DeclareCaptionStyle{ruled}{labelfont=normalfont,labelsep=colon,strut=off} 
\floatstyle{ruled}
\newfloat{listing}{tb}{lst}{}
\floatname{listing}{Listing}
\title{Diffusion Reconstruction-based Data Likelihood Estimation for Core-Set Selection}

\author{
    Mingyang Chen\textsuperscript{\rm 1,2},
    Jiawei Du\textsuperscript{\rm 3},
    Bo Huang\textsuperscript{\rm 1,2},
    Yi Wang\textsuperscript{\rm 4},
    Xiaobo Zhang\textsuperscript{\rm 5},
    Wei Wang\textsuperscript{\rm 1,2}
}

\affiliations{
    \textsuperscript{\rm 1}The Hong Kong University of Science and Technology (Guangzhou)\\
    \textsuperscript{\rm 2}The Hong Kong University of Science and Technology\\
    \textsuperscript{\rm 3}CFAR, A*STAR, Singapore
    \textsuperscript{\rm 4}Dongguan University of Technology
    \textsuperscript{\rm 5}Southwest Jiaotong University\\
    mchenbt@connect.ust.hk, weiwcs@ust.hk
}

\usepackage{bibentry}

\begin{document}

\maketitle

\begin{abstract}


Existing core-set selection methods predominantly rely on heuristic scoring signals such as training dynamics or model uncertainty, lacking explicit modeling of data likelihood. This omission may hinder the constructed subset from capturing subtle yet critical distributional structures that underpin effective model training.
In this work, we propose a novel, theoretically grounded approach that leverages diffusion models to estimate data likelihood via reconstruction deviation induced by partial reverse denoising. Specifically, we establish a formal connection between reconstruction error and data likelihood, grounded in the Evidence Lower Bound (ELBO) of Markovian diffusion processes, thereby enabling a principled, distribution-aware scoring criterion for data selection. Complementarily, we introduce an efficient information-theoretic method to identify the optimal reconstruction timestep, ensuring that the deviation provides a reliable signal indicative of underlying data likelihood. Extensive experiments on ImageNet demonstrate that reconstruction deviation offers an effective scoring criterion, consistently outperforming existing baselines across selection ratios, and closely matching full-data training using only 50\% of the data.  Further analysis shows that the likelihood-informed nature of our score reveals informative insights in data selection, shedding light on the interplay between data distributional characteristics and model learning preferences. The code is available at \url{https://github.com/mchen725/DRD}.

%
%

\end{abstract}

\section{Introduction}
\label{intro}

Data selection, particularly core-set selection, is crucial for enabling efficient and scalable training of deep neural networks by identifying representative subsets from massive datasets. The exponential growth of large-scale datasets across various domains~\cite{kolesnikov2020big,brown2020language,radford2021learning} has rendered the direct utilization of entire datasets computationally prohibitive. Effective core-set selection can significantly accelerate model training, enhance resource efficiency and provide insights into learning preferences \cite{paul2021deep,guo2022deepcore}. This is especially beneficial in resource-intensive downstream applications such as fine-tuning large-scale foundation models~\cite{joaquin2024in2core,xia2024less}, continual learning~\cite{DBLP:conf/iclr/YoonMYH22,NEURIPS2023_a0251e49}, and multi-task learning~\cite{DBLP:conf/emnlp/KungYCYC21,DBLP:conf/acl/RenduchintalaBR24}.

Existing core-set selection methods can generally be classified into score-based and optimization-based approaches \cite{DBLP:conf/icml/ChoiKC24}. Score-based methods, popular for their scalability, typically leverage heuristic signals such as model uncertainty~\cite{DBLP:conf/nips/Pleiss0EW20}, training dynamics including forgetting events \cite{DBLP:conf/iclr/TonevaSCTBG19}, and loss-based metrics like EL2N~\cite{DBLP:conf/nips/PaulGD21}. Despite the practical utility, these methods lack explicit modeling of data likelihood, leading such surrogate-based heuristic selections may fail to capture nuanced yet critical distributional characteristics intrinsic to model training. 
The comparison shown in Figure~\ref{fig:tsne_selection} further illustrates that these heuristic scores often fail to stratify samples based on their alignment with the underlying criterion, leading to selected subsets that resemble random sampling. This lack of sensitivity not only reduces their effectiveness and flexibility as selection criteria but also limits their interpretability in reflecting model learning preferences.

\begin{figure*}[t]
    \centering
    \includegraphics[width=\linewidth]{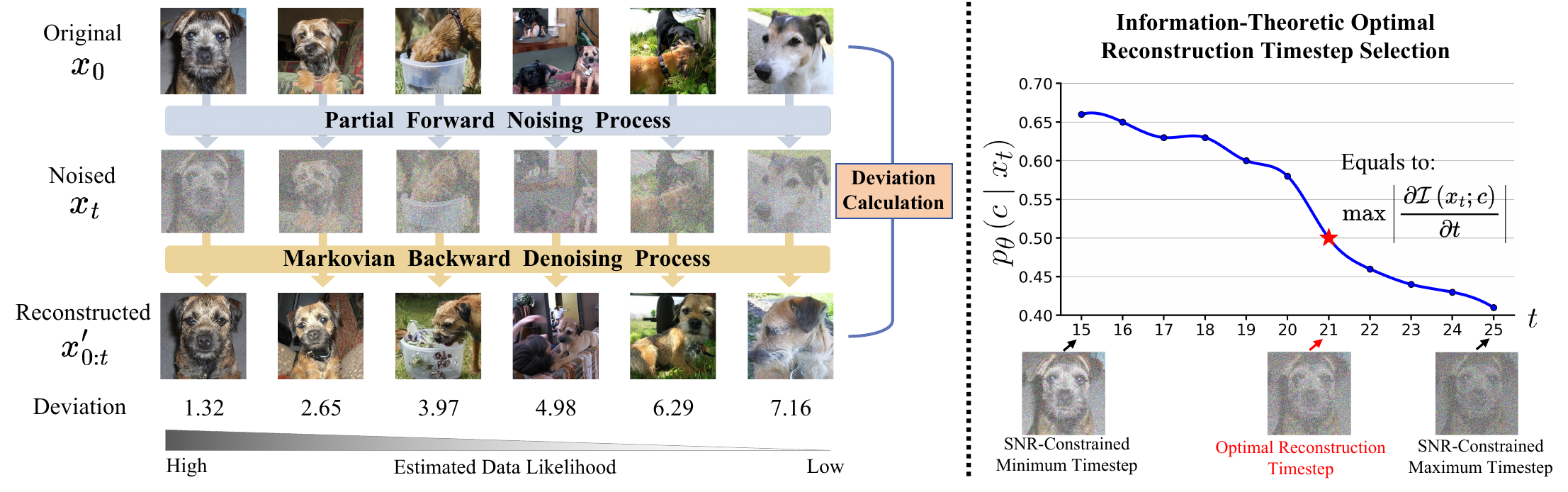}
    \caption{
    Intuitive illustration of our likelihood-informed scoring and optimal reconstruction timestep selection.
    \textbf{Left:} 
    The deviation between real data point \(x_0\) and the reconstructed \(x'_{0:t}\) serves as a likelihood-sensitive signal, with lower deviation indicating higher estimated data likelihood.
 The visualized examples reveal a semantic illustration: \textit{high-likelihood samples typically feature class-relevant objects that are spatially prominent and well-formed}; \textit{moderate-likelihood samples often contain target objects that are less visually salient}, e.g., occupying smaller regions or blended with irrelevant elements; and \textit{ low-likelihood samples exhibit apparent out-of-distribution characteristics}, leading to significant semantic shifts after reconstruction.
    \textbf{Right:} We select the optimal reconstruction timestep (\(0<t<T\)) by maximizing the drop rate \( \left| \partial \mathcal{I}(x_t; c)/\partial t \right| \). Following Lemma~\ref{lemma:mutual}, we equivalently maximize it by the time derivative of \( \log p_\theta(c \mid x_t) \) predicted by a \textit{diffusion classifier} \cite{DBLP:conf/iccv/LiPDBP23}. The search is constrained to \( \text{SNR}(t) \in [\gamma_{\min}, \gamma_{\max}] \) to avoid degenerate regions of timesteps.
    }
    \label{fig:intuitive}
\end{figure*}

To address this limitation, we propose a theoretically principled and distribution-aware data scoring criterion, grounded explicitly in diffusion generative models. Specifically, we exploit the inherent generative modeling capability of Denoising Diffusion Probabilistic Models (DDPMs) \cite{ho2020denoising} to directly quantify data likelihood through reconstruction deviation induced by partial reverse denoising, grounded in the Evidence Lower Bound (ELBO) of Markovian diffusion processes. Our core insight, illustrated intuitively in Figure~\ref{fig:intuitive}, is that the reconstruction error between real data and their partially denoised reconstructions provides a precise, likelihood-sensitive signal reflecting data distributional characteristics. Lower reconstruction deviations correspond to samples closely aligned with the authentic distribution, whereas higher errors indicate samples farther from high-probability regions.

A key aspect of our methodology involves strategically selecting the reconstruction timestep, as naively chosen timesteps either degenerate into trivial reconstructions at minimal noise levels or lose discriminative sensitivity at high noise levels. Inspired by Information Bottleneck (IB) theory~\cite{tishby2000information}, we formulate and solve this selection problem by maximizing the rate of decrease in mutual information between noised data and class labels, thus ensuring the most informative and discriminative timestep selection. This principled selection strategy avoids empirical grid search inefficiencies and theoretically grounds our approach in information-theoretic optimization.

In summary, our contributions are threefold:
\begin{itemize}
\item We present a novel, distribution-aware scoring criterion for core-set selection, grounded theoretically in the connection between diffusion reconstruction error and data likelihood.
\item We introduce an efficient information-theoretic approach to select reconstruction timesteps,  ensuring that the deviation provides a reliable signal indicative of underlying data likelihood.
\item Extensive experiments on ImageNet demonstrate that using our proposed reconstruction deviation as a scoring criterion consistently yields superior performance. Further analysis of the likelihood-informed nature of our method provides new insights into the relationship between model learning preferences and the distributional characteristics of training data.
\end{itemize}

\section{Background}


\paragraph{Diffusion Model Preliminaries.}
Denoising Diffusion Probabilistic Models (DDPMs)~\cite{ho2020denoising} define a generative process by learning to reverse a fixed forward noising process. Let \( x_0 \sim q^c(x) \) denote a data sample drawn from the real distribution conditioned on the class label \( c \in \{1, \dots, C\} \). The forward process gradually adds Gaussian noise over \( T \) discrete timesteps using a fixed variance schedule \( \{\beta_t\}_{t=1}^T \). The marginal distribution at step \( t \) can be expressed in closed form as:
\begin{equation}
    x_t = \sqrt{\bar{\alpha}_t} x_0 + \sqrt{1 - \bar{\alpha}_t} \, \epsilon, \quad \epsilon \sim \mathcal{N}(0, I),
\end{equation}
where \( \bar{\alpha}_t = \prod_{s=1}^t (1 - \beta_s) \) denotes the noise retention factor.
To generate data, diffusion models learn a reverse Markov process \( p_\theta(x_{t-1} \mid x_t, c) \) that approximates the time-reversed dynamics of the forward process. The model is trained by maximizing a variational lower bound (ELBO) on the conditional log-likelihood:
\begin{align}\label{eq:elbo}
    \log p_\theta(x_0 \mid c)
    &\geq
    \mathbb{E}_{q(x_{1:T} \mid x_0)} \left[ \log \frac{p_\theta(x_{0:T}, c)}{q(x_{1:T} \mid x_0)} \right] \notag \\
    &\approx
    -\mathbb{E}_{t, \epsilon} \left[ \left\| \epsilon - \epsilon_\theta(x_t, t, c) \right\|^2 \right] + \mathcal{C},
\end{align}
where \( \mathcal{C} \) is a constant independent of model parameters. In practice, \( p_\theta(x_{t-1} \mid x_t, c) \) is modeled as a Gaussian with its mean predicted by a neural network \( \epsilon_\theta(x_t, t, c) \) trained to reconstruct the injected noise \( \epsilon \).
At inference, one samples \( x_T \sim \mathcal{N}(0, I) \) and applies the learned reverse transitions:
\begin{equation}
    x_{t-1} = \frac{1}{\sqrt{\alpha_t}} \left( x_t - \frac{1 - \alpha_t}{\sqrt{1 - \bar{\alpha}_t}} \epsilon_\theta(x_t, t, c) \right) + \sigma_t z,\ z \sim \mathcal{N}(0, I),
\end{equation}
where \( \sigma_t^2 \) is the posterior variance and \( \alpha_t = 1 - \beta_t \).


\paragraph{Core-Set Selection.}
Recent advances in core-set selection can be broadly categorized into score-based and optimization-based approaches. Score-based methods evaluate the importance or utility of individual data via surrogate metrics. These include instance-wise influence estimation methods such as data shapley~\cite{DBLP:conf/icml/GhorbaniZ19,DBLP:conf/aistats/KwonRZ21,DBLP:conf/aistats/Kwon022},  influence functions \cite{DBLP:conf/icml/KohL17,pruthi2020estimating}; and training dynamics based metrics such as forgetting events \cite{DBLP:conf/iclr/TonevaSCTBG19}, EL2N \cite{DBLP:conf/nips/PaulGD21}, memorization \cite{DBLP:conf/nips/FeldmanZ20}, and CG-score \cite{DBLP:conf/iclr/KiCC23}. In contrast, optimization-based methods formulate subset selection as an optimization problem, aiming to approximate the diverse characteristics of the full dataset \cite{DBLP:conf/icml/MirzasoleimanBL20,DBLP:conf/icml/0007KM23,DBLP:conf/icml/PooladzandiDM22}. However, such methods are often computationally intensive and underperform compared to score-based ones \cite{DBLP:conf/icml/ChoiKC24}.
Our method also falls under the score-based category but differs fundamentally in both principle and implementation. Unlike prior approaches relying on certain surrogate behaviours,
our method derives selection signals from a theoretically grounded connection between diffusion reconstruction deviation and data likelihood, providing a distribution-aware criterion.
 To the best of our knowledge, this is the first work to leverage data likelihood as a scoring signal for core-set selection, offering a novel and interpretable perspective on the relationship between data distributional characteristics and model performance.

\paragraph{Reconstruction-based OOD Detection.}
Several recent works explored using diffusion models for out-of-distribution (OOD) detection by exploiting their reconstruction behaviour under partial noising or masking \cite{Graham_2023_CVPR,DBLP:conf/icml/LiuZWW23,DBLP:conf/cvpr/BellierA22}. These methods measure reconstruction errors at fixed or multiple noise levels as heuristics for identifying OOD data. While effective for OOD detection, they lack theoretical grounding in the data likelihood of in-distribution samples and provide no analysis for data selection. In particular, the choice of reconstruction timestep is typically made via empirical grid search. 
As there exists no principled metric to assess the discriminative utility for data selection, effectiveness can only be validated via downstream training, which is computationally expensive.
In contrast, our method establishes a formal connection between reconstruction deviation and data likelihood and proposes an efficient, information-theoretic criterion for selecting reconstruction timesteps aligned with reliable likelihood estimation.


\section{Method}
\label{method}

\subsection{Estimating Data Likelihood with Diffusion Reconstruction Deviation}

In this work, we propose to leverage diffusion generative models, which are explicitly trained to learn the authentic data distribution, as a natural lens through which data likelihood can be assessed. Specifically, we show that the deviation between a sample and its reconstruction via Markovian diffusion provides an effective, likelihood-sensitive signal for identifying samples that are most aligned with the selection criteria.

To formalize this idea, we consider the reconstruction deviation between a real data point \( x_0 \) and its denoised estimate \( x_{0:t}' \), obtained by reversing a partial forward process from an intermediate noised \( x_t \). This deviation is defined as \( \Delta x_0(t) := \left|x_{0:t}' - x_0\right| \).  The following theorem formalizes the inverse relationship between the reconstruction deviation and the log-likelihood of a data point.

\begin{theorem}[Inverse Dependence of Reconstruction Deviation on Log-Likelihood]
\label{thm:thm1}
Let \( x_0 \in \mathbb{R}^d \), \( x_t = \sqrt{\bar{\alpha}_t} x_0 + \sqrt{1 - \bar{\alpha}_t} \epsilon \) where \( \epsilon \sim \mathcal{N}(0, I) \), and \( x_{0:t}' \) be the reconstructed data obtained by DDPM denoising from \(x_t\),
the expected squared reconstruction deviation satisfies:
\[
\mathbb{E}_{\epsilon} \left[ \| \Delta x_0(t) \|^2 \right] \geq -\kappa(t) \log q(x_0) + \mathcal{C}_{\text{noise}}(t),
\]
where \( \kappa(t) = \frac{1}{t} \sum_{s=1}^t \frac{1}{\sigma_s^2} \), and \( \mathcal{C}_{\text{noise}}(t) \) is a constant independent of \(x_0\).
\end{theorem}


The detailed proof is provided in Appendix A.1. Intuitively, Theorem~\ref{thm:thm1} suggest that when \( x_0 \) resides in a high-density region, the reverse process should reconstruct it with relatively small distortion. In contrast, low-density or off-manifold samples tend to incur larger reconstruction errors.  While the theorem provides a lower bound rather than an exact equivalence, it establishes a principled, theoretically grounded link between reconstruction deviation and data likelihood, justifying the use of deviation as a distribution-aware scoring signal for core-set selection.

In practice, we adopt the Deterministic Denoising Implicit Model (DDIM)~\cite{song2021denoising} for reconstructing \(x'_{0:t}\), which preserves the same marginal distributions as DDPM under the given noise schedule while enabling more efficient denoising from intermediate steps \( x_t \).
The DDIM update is given by:
\begin{equation}
    x_{t-1} = \sqrt{\bar{\alpha}_{t-1}} \left( \frac{x_t - \sqrt{1 - \bar{\alpha}_t} \, \epsilon_\theta(x_t, t)}{\sqrt{\bar{\alpha}_t}} \right) + \sqrt{1 - \bar{\alpha}_{t-1}}  \epsilon.
\end{equation}

To robustly measure reconstruction deviation, we use the Learned Perceptual Image Patch Similarity (LPIPS), which computes perceptual similarity via the \( \ell_2 \) distance between deep features from a pretrained network (e.g., ResNet~\cite{DBLP:conf/cvpr/HeZRS16}), and is known to correlate well with human judgments~\cite{DBLP:conf/cvpr/ZhangIESW18}.

\subsection{Information-Theoretic Optimal Reconstruction Timestep Selection}
\label{sec:optimal_t}

The effectiveness of reconstruction deviation as a proxy for data likelihood critically depends on the choice of the target reconstruction timestep $t$. At extreme timesteps, the approach degenerates: when $t \to T$, the noised input $x_t$ approaches pure Gaussian noise, making the reconstruction independent of $x_0$; conversely, when $t \to 0$, minimal noise injection yields $x'_0 \approx x_0$, rendering the deviation uninformative. 
Consequently, an effective timestep $t$ should be selected to distinguish between high and low-likelihood data. However, identifying such an optimal \( t \) is challenging due to the absence of a principled metric for discriminative utility and the high cost of downstream validation.

\begin{figure*}[t]
    \centering
    \includegraphics[width=0.97\linewidth]{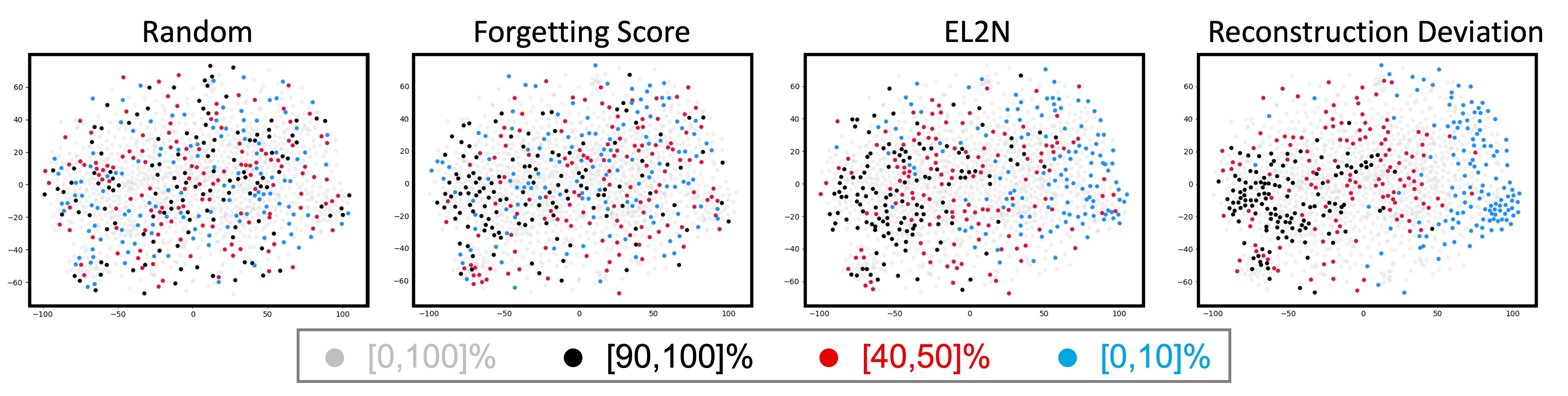}
    \caption{
    t-SNE visualization of stratified samples from the ImageWoof dataset. Samples are grouped by ascending score ranges for Forgetting Score~\cite{DBLP:conf/iclr/TonevaSCTBG19}, EL2N~\cite{DBLP:conf/nips/PaulGD21}, and our proposed Reconstruction Deviation, with Random representing equally sized random groups for reference. The distributions show that Forgetting Score and EL2N produce stratifications that resemble random sampling, whereas Reconstruction Deviation yields more distinct and semantically coherent groupings.
    }
    \label{fig:tsne_selection}
\end{figure*}

Intuitively, an optimal timestep $t$ should balance information preservation about the original data and discriminative sensitivity regarding label information. This criterion naturally aligns with minimizing an information bottleneck (IB) formulation \cite{tishby2000information} involving the original data $x_0$, intermediate noised data $x_t$, and associated class labels $c$:
\begin{equation}
    \min_{t} \mathbb{E}_{x_0}\left[\mathcal{I}(x_0; x_t) - \beta \mathcal{I}(x_t; c)\right],
\end{equation}
where $\mathcal{I}(\cdot ; \cdot)$ denotes mutual information, and $\beta > 0$ trades off between compression of the original information and preservation of discriminative power. In this IB context, $x_t$ is viewed as a compressed representation of $x_0$, regulated by the timestep $t$, and $\mathcal{I}(x_0; x_t)$ quantifies the retained information content. Thus, this quantity can be approximately characterized using the signal-to-noise ratio $\text{SNR}(t) = \bar{\alpha}_t / (1 - \bar{\alpha}_t)$ \cite{DBLP:journals/corr/abs-2208-11970}. To circumvent the instability and computational overhead associated with the selection of the hyperparameter $\beta$, we propose using $\mathcal{I}(x_0; x_t)$ as a regularization criterion. Specifically, we constrain the candidate range of timesteps $t$ by restricting the signal-to-noise ratio $\text{SNR}(t)$ to lie within the empirical interval $[ \gamma_{\min}, \gamma_{\max} ] = [0.05, 1]$. 

Conversely, $\mathcal{I}(x_t; c)$ quantifies the mutual information between the intermediate noised data $x_t$ and its corresponding class $c$. As $t$ increases, this term decreases monotonically and asymptotically approaches zero. Consequently, directly maximizing $\mathcal{I}(x_t; c)$ leads to the selection of $t$ at the upper bound $\text{SNR}(t) = \gamma_{\max}$, degenerating the selection strategy into a naive grid search. Intuitively, the optimal timestep $t$ should correspond to the boundary at which the distribution $q(x_t | x_0)$ transitions significantly from authentic data towards noise. At this boundary, the rate of decrease of mutual information between $x_t$ and labels $c$ should be maximized. Based on this insight, we formally define the optimal timestep selection criterion as:
\begin{equation}
t^* = \arg\max_{t} \left| \frac{\partial \mathcal{I}(x_t; c)}{\partial t} \right| \ \text{s. t.} \ \text{SNR}(t) \in [\gamma_{\min}, \gamma_{\max}].
\end{equation}

\begin{lemma}\label{lemma:mutual}
    Let \( x_t = \sqrt{\bar{\alpha}_t} x_0 + \sqrt{1-\bar{\alpha}_t} \epsilon \), where \( \epsilon \sim \mathcal{N}(0, I) \), and assume a uniform prior over classes \( p(c) = \frac{1}{C} \). For all \( t \in [0, T] \),  
\[
    \left|\frac{\partial \mathcal{I}(x_t; c)}{\partial t}\right| = \left|\mathbb{E}_{x_0, c}\left[\frac{\partial}{\partial t}\log p(c\mid x_t)\right]\right|.
\]
\end{lemma}

The detailed proof of Lemma \ref{lemma:mutual} is provided in Appendix A.2. According to this lemma, the time derivative of mutual information can alternatively be computed through the expectation of the time derivative of \( \log p(c \mid x_t) \). Based on Bayes' theorem and the  uniform prior assumption of \( p(c) \), we employ the diffusion classifier \cite{DBLP:conf/iccv/LiPDBP23} to approximate class probabilities based on the ELBO formulation of the data log-likelihood shown in Equation (\ref{eq:elbo}):
\begin{equation}\label{eq:pc}
    p_\theta(c \mid x_t) = \frac{\exp\left\{-\mathbb{E}_{\epsilon}[\| \epsilon - \epsilon_\theta(x_t, c) \|^2]\right\}}{\sum_{c' \in C} \exp\left\{-\mathbb{E}_{\epsilon}[\| \epsilon - \epsilon_\theta(x_t, c') \|^2]\right\}}.
\end{equation}

Note that we omit the expectation over \( t \) as discussed in \cite{DBLP:conf/iccv/LiPDBP23} for testing on each candidate timestep. Furthermore, since the input timestep \( t \) is inherently discrete within practical diffusion formulation, the partial derivative \( \frac{\partial}{\partial t} \log p_\theta(c \mid x_t) \) cannot be directly calculated using automatic differentiation tools (e.g., PyTorch's autograd). To address this limitation, we propose using a finite difference method to approximate the partial derivative:
\begin{equation}
    \frac{\partial}{\partial t} \log p_\theta(c \mid x_t) \approx \frac{\log p_\theta(c \mid x_{t + \Delta t}) - \log p_\theta(c \mid x_{t - \Delta t})}{2 \Delta t},
\end{equation}
where \( \Delta t = 1 \) in our practical implementation.
Finally, we perform a Monte Carlo estimate of the expectation over $x_0 \sim q^c(x)$.
Specifically, for each class in the target dataset, we randomly sample \( B \) data points from its class-conditional subset. The class-wise optimal reconstruction timestep is determined by solving the following objective:
\begin{equation}\label{eq:tc*}
\begin{gathered}
    t^*_c = \arg\max_t\; \frac{1}{B} \sum_{i=1}^{B} \left| 
    \frac{\log p_\theta(c \mid x^{(i)}_{t + \Delta t}) - \log p_\theta(c \mid x^{(i)}_{t - \Delta t})}{2 \Delta t} 
    \right| \\
    \text{subject to}\quad \text{SNR}(t) \in [\gamma_{\min}, \gamma_{\max}].
\end{gathered}
\end{equation}

\subsection{Analysis of Likelihood-Informed Core-Set Construction}
\label{sec:analysis}

\paragraph{Better Stratification and Interpretability of Distribution-Aware Likelihood Score.}
The foundation of data  selection lies in the assumption that the score function quantifies how well each sample aligns with a target property, such as forgetting events, model uncertainty, or, in our case, distributional fidelity. Thus, an informative score should induce a meaningful stratification of the dataset: samples across different score quantiles ought to exhibit distinct structural distributions. In contrast, ineffective scores tend to produce subsets whose distributions resemble random sampling.
In Figure~\ref{fig:tsne_selection}, we visualize t-SNE embeddings of samples selected by different scoring methods. Compared to EL2N and Forgetting Score, which yield noticeably entangled quantile regions resembling quasi-random partitions, our reconstruction deviation score produces clearer separation across quantiles.
Interestingly, we also observe that the dense cluster of black points in the left-central region of the Forgetting plot, corresponding to the most frequently forgotten samples, aligns most closely with our stratification. The likelihood-informed nature of our score suggests that the Forgetting Score tends to favor low-likelihood data while lacking discriminative ability among less-forgotten examples. 

\begin{figure}[t] %
  \centering
  \includegraphics[width=0.43\textwidth]{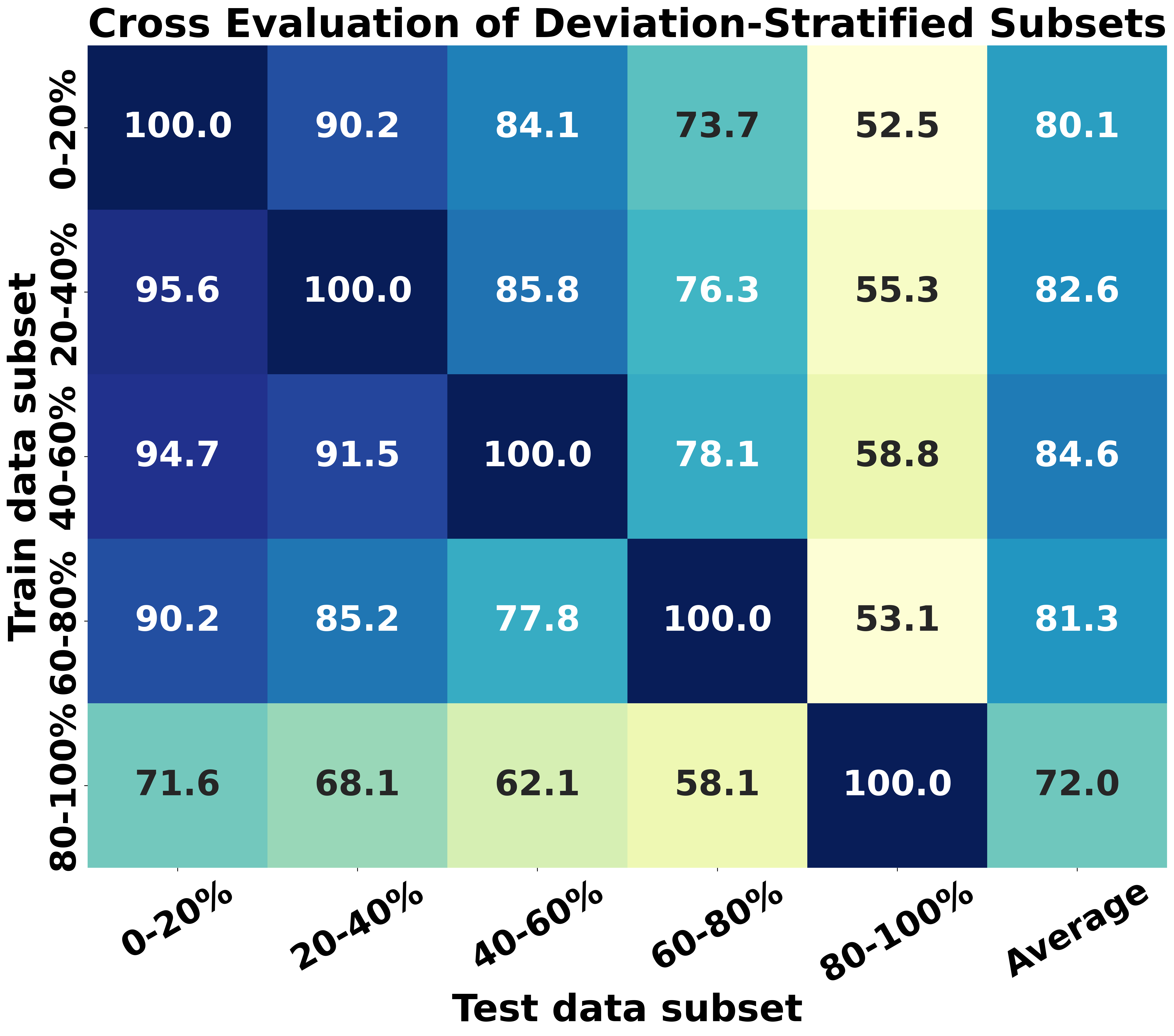} %
  \caption{Cross evaluation of ResNet-18 models trained on deviation-stratified subsets. Each model is trained on one subset and tested across all five. Lower indices (e.g., \(0-20\%\)) indicate higher estimated likelihoods.}
  \vspace{-8pt}
  \label{fig:train_split}
\end{figure}

\paragraph{Moderate-Likelihood Data Benefits Core-Set Construction.}
A natural follow-up question to our earlier findings is: \textit{which regions of the likelihood spectrum yield the most informative subset for model training?} To explore this, we adapt the “training set split” experiment from~\cite{DBLP:conf/icml/ChoiKC24}. Specifically, we sort the ImageWoof training set by reconstruction deviation and divide it into five equal-sized subsets, from the top 20\% (highest estimated likelihood) to the bottom 20\%. We train a ResNet-18 model on each subset and evaluate its accuracy across all five. As shown in Figure~\ref{fig:train_split}, models trained on moderate-likelihood subsets consistently generalize better, whereas those trained solely on the lowest or highest extremes perform poorly. 
To deepen understanding, we evaluate two contrasting selection strategies for leveraging our reconstruction-based score. \textbf{(1) Coverage-Centric Selection (CCS)}~\cite{zheng2023coveragecentric} divides per-class samples—sorted by reconstruction deviation—into \(\mathcal{B}\) strata and uniformly samples from each until the budget is met. \textbf{(2) Best Window Selection (BWS)}~\cite{DBLP:conf/icml/ChoiKC24} scans fixed-width windows over the sorted list using a step size of 5\%, with starting points in the range \([0, \min(50\%, 100\%-budget)]\), and selects the one achieving the highest validation accuracy on the full set.
As shown in Table~\ref{tab:ccs_vs_bws}, BWS consistently outperforms CCS across different selection ratios. Notably, we observe that, except at extremely low or high ratios, the optimal window start point generally falls within the \([20\%, 40\%]\) range. We attribute this to BWS selecting moderate-likelihood subsets that better align with the curriculum-like preference in model training \cite{DBLP:conf/icml/HacohenW19}, whereas CCS may introduce overly heterogeneous features, especially under tight budgets.
Moreover, the model's preference for moderate-likelihood data may arise from a similar effect as data augmentation techniques such as CutMix~\cite{DBLP:conf/iccv/YunHCOYC19}, which improve robustness by introducing mixed or ambiguous class signals. As illustrated in Figure~\ref{fig:intuitive} and 7-8 in Appendix E, moderate-likelihood samples often contain less salient features, thereby prompting the model to generalize under partial or uncertain semantic cues.



\section{Experiments}\label{sec:exp}

\subsection{Experimental Setup}

\paragraph{Evaluation Protocol.} 
We conduct core-set selection experiments on three datasets with increasing levels of difficulty: ImageNette, ImageWoof, and ImageNet-1K~\cite{DBLP:journals/ijcv/RussakovskyDSKS15}. 
ImageNette contains 10 classes with low intra-class similarity, while ImageWoof comprises 10 visually similar dog breeds, posing a more challenging classification task. 
We use ResNet-18 for ImageNette and ImageWoof, and ResNet-50 for ImageNet-1K. 
We compare our \textbf{D}iffusion \textbf{R}econstruction \textbf{D}eviation (\textbf{DRD}) method against seven baselines, including: 
(1) \textbf{Random}, 
(2) \textbf{Forgetting}~\cite{DBLP:conf/iclr/TonevaSCTBG19}, 
(3) \textbf{EL2N}~\cite{DBLP:conf/nips/PaulGD21}, 
(4) \textbf{AUM}~\cite{DBLP:conf/nips/Pleiss0EW20}, 
(5) \textbf{Moderate}~\cite{DBLP:conf/iclr/XiaL0S0L23}, 
and two score-oriented selection strategies—(6) \textbf{CCS}~\cite{zheng2023coveragecentric} and (7) \textbf{BWS}~\cite{DBLP:conf/icml/ChoiKC24}. Forgetting score is used as the underlying criterion for both CCS and BWS, following their default setup.
More details about the baselines and experiments are provided in Appendix B. 

\begin{figure*}[t]
    \centering
    \includegraphics[width=\linewidth]{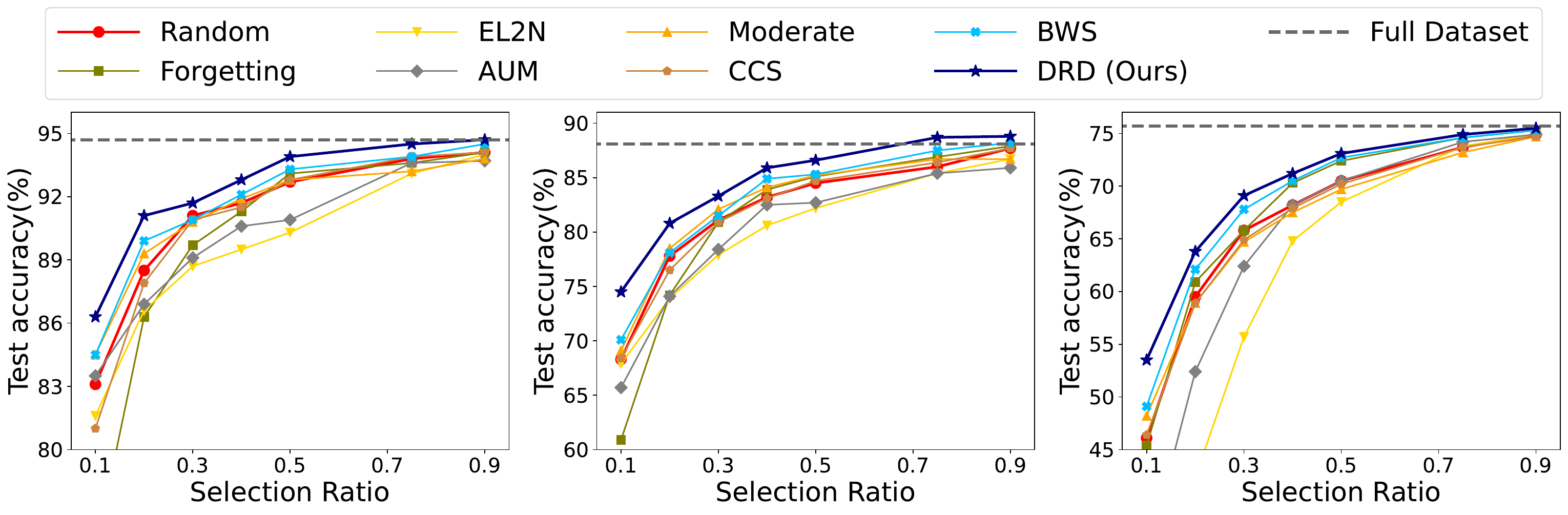}
        
    \vspace{- 0.5em}
    
    \caption*{
    \footnotesize
        \hspace{0.01\textwidth}(a) ImageNette
        \hspace{0.19\textwidth}(b) ImageWoof
        \hspace{0.19\textwidth}(c) ImageNet-1K
    }
    \caption{
        \textbf{(a, b, c) Core-set selection results.} 
        Test accuracy of models trained on data subsets selected by different methods at varying selection ratios on ImageNette, ImageWoof, and ImageNet-1K. 
        Our DRD method outperforms all baselines across selection budgets and approaches full-data performance with only 50\% of the data.
        See Appendix C for detailed results.
    }
    \label{fig:main_results}
\end{figure*}

\paragraph{Implementation Details.} 
We use the pretrained Diffusion Transformer (DiT) model from the official PyTorch implementation\footnote{\url{https://github.com/facebookresearch/DiT}}, trained with a full diffusion schedule of \( T = 1000 \) steps. During inference, we adopt the standard DDIM sampler with \(T=50\) and perform partial-step reconstruction from a selected \( t < 50 \), determined by our IB-based timestep selection strategy. To approximate \( p_{\theta}(c \mid x_t) \) in Equation~(\ref{eq:pc}), we sample 20 groups of random noise. We set \( B = 20 \) for the Monte Carlo estimate in Equation~(\ref{eq:tc*}). For data selection, we use the BWS strategy by default, scanning fixed-width windows over the  samples sorted by DRD score and selecting the one yielding the highest validation accuracy on the full training set. Detailed window start points are reported in Appendix B.2. All reported results of our method can be obtained using a single RTX 4090 GPU.

\subsection{Benchmark Evaluations}

In Figure~\ref{fig:main_results}, we report the test accuracy of models trained on data subsets selected by various methods across a range of selection ratios on ImageNette, ImageWoof, and ImageNet-1K. Across all benchmarks and selection budgets, our proposed DRD score consistently achieves superior performance. 
Notably, DRD achieves near full-data performance using only 50\% of the data across all datasets.
A particularly informative comparison lies in contrasting our DRD method with BWS under a shared selection strategy. Both methods adopt an identical sliding-window mechanism guided by full-data validation accuracy to identify optimal windows. The only difference stemming from the scoring criterion is that BWS defaults to using the Forgetting score, whereas DRD relies on our likelihood-sensitive reconstruction deviation. Despite operating under the same selection protocol, DRD achieves significantly better results, underscoring the effectiveness of our score. Specifically, under selection ratios \(\leq 40\%\), DRD surpasses BWS (with Forgetting score) by average margins of 1.3\%, 2.5\%, and 2.0\% on ImageNette, ImageWoof, and ImageNet-1K, respectively.

Furthermore, we provide a cross-architecture evaluation in Appendix D, where different data selection methods are evaluated on EfficientNet-B0 and ViT across two datasets. Overall, DRD generally achieves the highest test accuracy under different selection ratios, with particularly notable gains on the more challenging ImageWoof dataset, demonstrating its strong cross-architecture generalizability.

\begin{table*}[h]
\begin{minipage}{0.48\textwidth}
\centering
\small
\captionof{table}{\textbf{Score-oriented selection strategy comparison.} Test accuracy on  ImageWoof under varying selection ratios using two selection strategies: CCS (bin-based coverage-centric sampling) and BWS (validation-guided window search), each applied to three scores.  }
\begin{tabular}{c|c|ccc} \toprule
Selection            & Score      & 10\%              & 30\%              & 75\%              \\ \midrule
\multirow{3}{*}{CCS} & Forgetting & 68.4±0.9          & 80.9±0.3          & 86.4±0.2          \\
                     & EL2N       & 64.1±0.5          & 75.8±0.6          & 83.6±0.8          \\
                     & DRD        & 68.9±0.1          & 80.1±0.8          & 86.9±0.5          \\ \midrule
\multirow{3}{*}{BWS} & Forgetting & 70.1±0.7          & \textit{81.5±0.5} & 87.3±0.8          \\
                     & EL2N       & 68.9±0.5          & 80.9±0.5          & 86.5±0.3          \\
                     & DRD        & \textbf{74.5±0.5} & \textbf{83.3±0.4} & \textbf{88.7±0.8} \\ \bottomrule
\end{tabular}\label{tab:ccs_vs_bws}
\end{minipage}
\hspace{1.5em}
\begin{minipage}{0.45\textwidth}
\centering
\small
\captionof{table}{\textbf{Hyperparameter sensitivity of timestep selection.} Test accuracy on ImageWoof under varying selection ratios with different numbers of class-conditional samples \(B\) and noise samples \(\#\epsilon\) used in our  IB-based reconstruction timestep selection strategy. 
}
\begin{tabular}{c|c|ccc} \toprule
$B$               & $\#\ \epsilon $ & 10\%              & 30\%              & 75\%              \\ \midrule
\multirow{3}{*}{20} & 5              & 67.8±0.4          & 81.1±0.6          & 86.3±0.2          \\
                    & 20             & 74.5±0.5          & 83.3±0.4          & 88.5±0.8          \\
                    & 40             & 74.6±0.4          & 83.1±0.4          & 88.6±0.7          \\ \midrule
\multirow{3}{*}{40} & 5              & 68.3±0.4          & 82.1±0.5          & 87.1±0.4          \\
                    & 20             & \textbf{74.7±0.3} & 83.3±0.4          & 88.5±0.6          \\ 
                    & 40             & 74.6±0.3          & \textbf{83.4±0.4} & \textbf{88.7±0.7} \\ \bottomrule
\end{tabular}\label{tab:hyper}
\end{minipage}
\end{table*}


\subsection{Ablation Study and Analysis}\label{sec:ablation}

\paragraph{Score-Oriented Selection Strategy Comparison.}

To better understand the impact of selection strategy on overall performance, we compare two score-oriented strategies CCS and BWS across three scoring functions: Forgetting, EL2N, and our  DRD. Experiments are conducted on ImageWoof under varying selection ratios. As shown in Table~\ref{tab:ccs_vs_bws}, BWS consistently outperforms CCS across all scoring functions and selection ratios, highlighting the advantage of window-based selection over bin-based coverage. 
Among the three scores, DRD basically achieves the best performance under both strategies, confirming its strength as a likelihood-sensitive and distribution-aware signal. Notably, BWS with DRD yields the highest accuracy across all selection ratios. These results suggest that DRD not only provides a more effective ranking signal but also synergizes particularly well with curriculum-aligned strategies such as BWS.

\begin{figure*}[htbp]
	\centering
	\begin{minipage}[t]{0.48\textwidth}
		\centering
		\includegraphics[width=\linewidth]{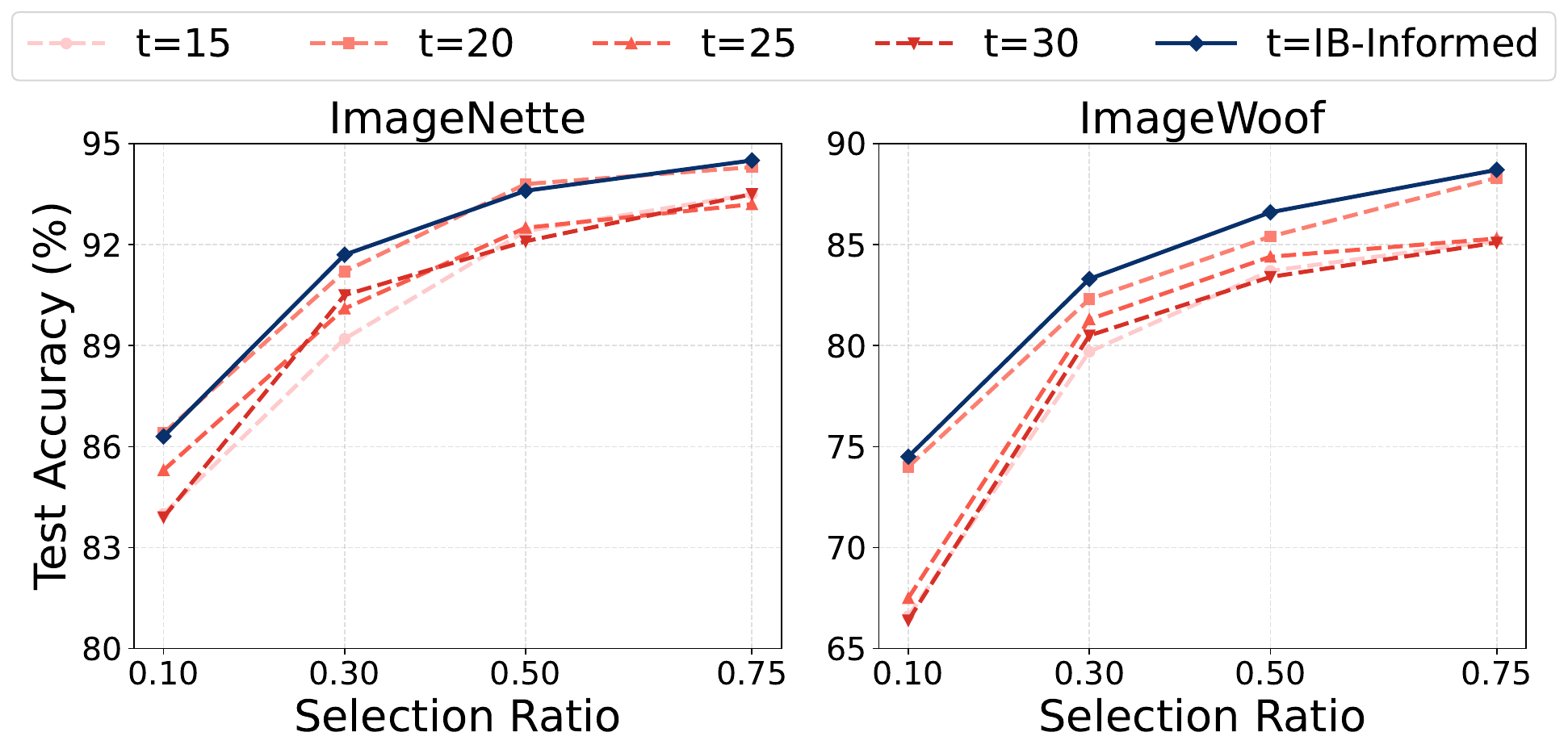}
		\caption{\textbf{Comparison of different reconstruction timesteps.} Test results of fixed timesteps and timesteps selected by our IB-informed method on ImageNette and ImageWoof. 
			While grid search requires full reconstruction and evaluation over the entire dataset, our method identifies effective, class-wise timesteps using lightweight Monte Carlo estimates, \textit{enabling timestep selection within minutes}.}
		\label{fig:timestep}
	\end{minipage}
	\hfill
	\begin{minipage}[t]{0.47\textwidth}
		\centering
		\includegraphics[width=\linewidth]{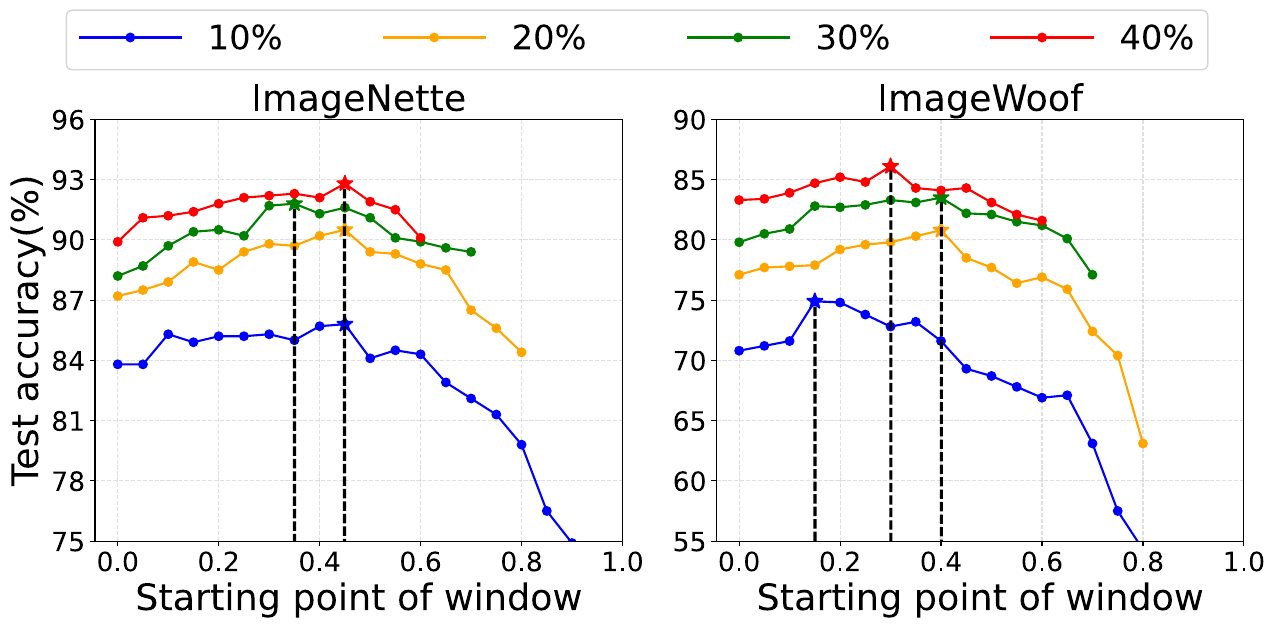}
		\caption{ \textbf{Comparison of selected window start points based on DRD score.} Test accuracy on ImageNette and ImageWoof when sliding a fixed-width selection window across the DRD-sorted score list. 
			High-performing subsets consistently arise from windows starting between 20\% and 40\% of the ranked list, reflecting a moderate-likelihood preference in model learning.}
		\label{fig:window_start}
	\end{minipage}
\end{figure*}

\paragraph{Effectiveness of Information-Theoretic Timestep Selection.}
Figure~\ref{fig:timestep} compares the performance of our information-bottleneck-informed (IB-informed) timestep selection method against fixed timesteps chosen via naive grid search on ImageNette and ImageWoof. Across all selection ratios, our method generally yields superior test accuracy, demonstrating that the class-wise  deviations computed using our selected timesteps lead to more discriminative likelihood estimation.
Empirically, the selected timesteps tend to cluster around \(t = 20\), explaining why fixed \(t=20\) performs comparably well. This suggests that the noise level at this range strikes a balance between retaining class-relevant features and enabling informative reconstruction.
However, naive grid search requires computing reconstruction deviations for the entire dataset at multiple candidate timesteps, followed by data selection and full training for each setting, making it computationally prohibitive. In contrast, we identify effective timesteps using a lightweight Monte Carlo estimate over a small subset of samples, typically within a few minutes. 
This efficiency, together with its strong empirical performance, makes our approach both practical and principled for the core-set construction.



\paragraph{Hyperparameter Sensitivity of Reconstruction Timestep Selection.}
We conduct a sensitivity analysis on two key hyperparameters of our information-theoretic timestep selection strategy: the number of class samples \(B\) used for Monte Carlo estimation in Equation~(\ref{eq:tc*}), and the number of noise samples \(\#\epsilon\) used to approximate \(p_{\theta}(c \mid x_t)\) of the diffusion classifier as defined in Equation~(\ref{eq:pc}). Results on ImageWoof are summarized in Table~\ref{tab:hyper}.
We observe that using an extremely small number of noise samples (e.g., \(\#\epsilon = 5\)) leads to noticeably degraded performance, particularly under low selection ratios. This is likely due to unreliable estimation of \(p_{\theta}(c \mid x_t)\), which in turn causes instability in the estimated timestep derivative and, ultimately, degrades the reliability of the reconstruction deviation. However, once \(\#\epsilon\) is set to 20 or higher, performance stabilizes and becomes largely insensitive to further increases. A similar trend is observed with respect to \(B\).
Based on these findings, we set \(B = 20\) and \(\#\epsilon = 20\) as default values. 

\paragraph{Analysis of Selection Window Start Points.}

To investigate how the structure induced by our DRD score relates to core-set quality, we perform a sliding-window analysis over the sorted score list. For each selection ratio, we fix the window width and vary its starting point, then evaluate the test accuracy of models trained on the resulting subsets. The results, shown in Figure~\ref{fig:window_start}, reveal consistent and interpretable patterns across both ImageNette and ImageWoof.
In particular, we find that the highest-performing subsets consistently emerge from a mid-range window—typically starting between the 20\% and 40\% mark in the DRD-sorted list. Performance degrades sharply when selecting from either extreme, suggesting that samples with the highest or lowest DRD scores contribute less effectively to learning. This observation reinforces our earlier hypothesis: the most informative and discriminative subsets often lie within the moderate-likelihood region, where examples are neither trivially easy nor overly noisy.


\section{Conclusion}\label{sec:conclusion}


In this work, we propose a novel and principled approach to core-set selection by leveraging diffusion-based reconstruction deviation as a likelihood-informed scoring criterion. By formally linking reconstruction error to data likelihood through the ELBO of diffusion models and guided by an information-theoretic reconstruction timestep selection strategy, our method offers both theoretical interpretability and practical effectiveness. Experiments on ImageNet demonstrate consistent performance gains over existing score-based methods. Besides empirical improvements, our analysis sheds light on the relationship between data distributional characteristics and model learning preferences.


\section {Acknowledgments}

Jiawei Du was supported by the A*STAR Career Development Fund (Grant No. C233312004) and by the National Research Foundation, Singapore
under its Digital Trust Centre Innovation Grant (DTC Award No:
DTC-IGC-02). 
Yi Wang was supported in part by the Guangdong Basic and Applied Basic Research Foundation (Grant No. 2023B1515120058). 
Wei Wang was supported by Advanced Materials-National Science and Technology Major Project (Grant No. 2025ZD0620100), Guangdong Provincial Key Laboratory of Integrated Communication, Sensing, and Computation for Ubiquitous Internet of Things (Grant No. 2023B1212010007), the Guangzhou Municipal Science and Technology Project (Grant Nos. 2023A03J0003, 2023A03J0013, and 2024A03J0621), and the Institute of Education Innovation and Practice Project (Grant No. HKUST(GZ)-ROP2025015).


\bibliography{aaai2026}


\end{document}